\pdfoutput=1

\documentclass[11pt]{article}

\usepackage{acl}

\usepackage{times}
\usepackage{latexsym}
\usepackage[most]{tcolorbox}
\usepackage{enumitem}

\usepackage[T1]{fontenc}

\usepackage[utf8]{inputenc}

\usepackage{microtype}

\usepackage{inconsolata}

\usepackage{graphicx}

\usepackage{hyperref}

\usepackage{amsmath}
\usepackage{amssymb}
\usepackage{booktabs}
\usepackage{balance}
\usepackage{listings}
\usepackage{xcolor}       
\usepackage{multirow}

\usepackage{amsmath, amsthm, amssymb}

\colorlet{lightyellow}{yellow!40}
\definecolor{navyblue}{rgb}{0.0, 0.0, 0.5}

\colorlet{lightyellow}{yellow!40}

\makeatletter
{\small 
\xdef\f@size@small{\f@size}
\xdef\f@baselineskip@small{\f@baselineskip}
\normalsize 
\xdef\f@size@normalsize{\f@size}
\xdef\f@baselineskip@normalsize{\f@baselineskip}
}
\newcommand{\smalltonormalsize}{%
  \fontsize
    {\fpeval{(\f@size@small+\f@size@normalsize)/2}}
    {\fpeval{(\f@baselineskip@small+\f@baselineskip@normalsize)/2}}%
  \selectfont
}
\makeatother

\newcommand{\myparagraph}[1]{\paragraph{#1}\mbox{}\\}

\title{MalAlgoQA: Pedagogical Evaluation of Counterfactual Reasoning in \\ Large Language Models and Implications for AI in Education\\}

\author{
  Naiming Liu$^*$\\
  Rice University \\
  \texttt{nl35@rice.edu} \\
  \And
  Shashank Sonkar$^*$\\
  Rice University \\
  \texttt{ss164@rice.edu} \\
  \AND 
  MyCo Le\\
  Rice University \\
  \texttt{msl12@rice.edu} \\
  \And
  Richard G. Baraniuk \\
  Rice University \\
  \texttt{richb@rice.edu} \\
  \\
}

\begin{document}

\maketitle
\begin{abstract}

This paper introduces MalAlgoQA, a novel dataset designed to evaluate the counterfactual reasoning capabilities of Large Language Models (LLMs) through a pedagogical approach.
The dataset comprises mathematics and reading comprehension questions, each accompanied by four answer choices and their corresponding rationales. 
At the heart of MalAlgoQA are ``malgorithms'' - rationales behind incorrect answer choices that represent flawed yet logically coherent reasoning paths.
These malgorithms serve as counterfactual scenarios, allowing us to assess an LLM's ability to identify and analyze flawed reasoning patterns.
We propose the Malgorithm Identification task, where LLMs are assessed based on their ability to identify corresponding malgorithm given an incorrect answer choice. 
To evaluate the model performance, we introduce two metrics: Algorithm Identification Accuracy (AIA) for correct answer rationale identification, and Malgorithm Identification Accuracy (MIA) for incorrect answer rationale identification.
Our experiments reveal that state-of-the-art LLMs exhibit significant performance drops in MIA compared to AIA, highlighting the challenges in counterfactual reasoning.
Surprisingly, we find that the chain-of-thought prompting technique not only fails to consistently enhance MIA but can sometimes lead to underperformance compared to simple prompting.
These findings have important implications for developing LLMs with improved counterfactual reasoning, particularly relevant for AI-powered tutoring systems, where identifying and addressing student misconceptions is essential. 
MalAlgoQA dataset is available  \href{https://github.com/luffycodes/MalAlgoQA-Dataset}{here}.

\end{abstract}

\renewcommand{\thefootnote}{}
\footnotetext{$^*$Equal contribution.}

\section{Introduction}

Evaluating the counterfactual reasoning capabilities of Large Language Models (LLMs) -- their ability to consider ``what if'' scenarios and infer causes based on hypothetical situations -- remains a significant challenge despite their impressive performance in various natural language tasks \citep{pearl2019seven,rafetseder2021counterfactual,dupe}.
We introduce a novel method to evaluate these capabilities, drawing inspiration from educational assessment techniques.
Specifically, we leverage the practice of distractor choice generation, which involves creating plausible but incorrect answer options by envisioning hypothetical scenarios and logical yet flawed reasoning paths. This approach mimics how educators anticipate and model potential student misconceptions, engaging in counterfactual thinking about alternative cognitive processes that could lead to specific errors.

\begin{table*}[ht!]
\centering
\begin{tcolorbox}[colback=gray!10, colframe=gray!80!black,title=\textbf{\large Question}]
\large Each side of a square is 3 inches long. A student increases the length of each side by 3 inches. The area of the new square is how many times the area of the original square?
\end{tcolorbox}

\begin{tcolorbox}[colback=navyblue!10,colframe=navyblue,title=\textbf{\large Answer Choices}]

\begin{enumerate}[label=(\Alph*)]
\item 9 (\normalsize Rationale: Squared the change in the length.)
\item 2 (\normalsize Rationale: Calculated the relationship between the sides.)
\item 3 (\normalsize Rationale: Used the change in the length.)
\item 4 (\normalsize Rationale: Calculated the area of the original square: $3 \times 3 = 9$ and the area of the new square with each side length increased by 3 inches: $6 \times 6 = 36$. Then determined that the area of the new square is 4 times the area of the original square.)
\end{enumerate}
\end{tcolorbox}

\caption{An example question from the MalAlgoQA dataset illustrating the Malgorithm Identification task. Each answer choice is associated with a rationale, representing the reasoning process that led to that answer. The task for the model is to correctly identify the rationale given a particular answer choice. For incorrect answer choices, the corresponding rationales are malgorithms, representing flawed reasoning processes and thereby evaluating the counterfactual reasoning abilities of LLMs.}
\label{tab:math_example}
\end{table*}

Building on this foundation, we present MalAlgoQA, a dataset designed to challenge LLMs' ability to reason about hypothetical situations and identify flawed logical pathways.
MalAlgoQA consists of a diverse set of 807 mathematics and 290 reading comprehension questions spanning various grade levels (3-11), content classifications (e.g., algebra, geometry, number \& operations), and Depth of Knowledge levels (DOK) (1-3) \citep{webb2002depth}. Each question is accompanied by a set of answer choices and their associated rationales.

Central to our approach is the concept of ``\textbf{malgorithms}'' - a term we coin to describe the flawed reasoning paths that lead to incorrect answers (name is inspired by the concept of ``mal-rules'' proposed by \citet{payne1990algebra}).
These malgorithms represent deterministic yet erroneous thought processes, mirroring the way students might logically follow a series of steps based on a misunderstanding or misapplication of concepts. We propose the \textbf{Malgorithm Identification task} to assess LLMs' ability to reason about both correct and incorrect answer rationales. In this task, given a question and a specific answer choice, the model must identify the underlying rationale (algorithm or malgorithm) that led to that choice. This approach challenges LLMs to engage in both causal and counterfactual reasoning. To illustrate, consider a simple question: ``What is 1 + 2 × 3 + 4?'' with choices A. 11, B. 13, and C. 21. The rationales are: ``Worked left to right'' (corresponding to choice B), ``Applied PEASMD rule instead of PEMDAS'' (corresponding to choice C), and ``Applied PEMDAS rule'' (corresponding to the correct choice A). Given an answer choice, the model should identify the rationale explaining that choice.

In order to evaluate model performance, we introduce \textbf{two key metrics: Algorithm Identification Accuracy (AIA)}, which measures accuracy in identifying the rationale behind a given correct answer. \textbf{Malgorithm Identification Accuracy (MIA)}, of particular interest, evaluates the model's counterfactual causal reasoning abilities by quantifying accuracy in identifying the rationale behind an incorrect answer. Our experiments reveal that GPT-4o \citep{openai2023gpt4} achieves an AIA of $95.7\%$, demonstrating its proficiency in this task. However, when it comes to MIA, GPT-4o's performance drops significantly, with an accuracy of only $66.1\%$. This pattern, where accuracy is higher for correct answer rationales and lower for incorrect answer rationales, is consistent across other language models, such as LLaMA-3-70B \citep{llama3modelcard} and GPT-3.5 \citep{schulman2022chatgpt}. It emphasizes the challenge of counterfactual reasoning for LLMs, especially on finding the underlying rationale behind incorrect answer choices. Our experiments also reveal an unexpected result that Chain-of-Thought (CoT) prompting \cite{cot,letsthinkstepbystep} not only fails to consistently enhance MIA, it sometimes even underperforms compared to simple prompting. 


The \textbf{implications} of our findings reveal critical challenges for \textbf{AI in education}. The stark performance gap between AIA and MIA across all models (e.g., GPT-4o's $95.7\%$ AIA vs. $66.1\%$ MIA) exposes fundamental limitations in LLMs' ability to understand student misconceptions. This discrepancy suggests that current \textbf{LLM tutoring systems fail to identify and address the root causes of student errors}.
 Such limitations necessitate a rethinking of feedback mechanisms in LLM-powered educational tools. 
 Their struggle with malgorithm identification (e.g., GPT-3.5's dramatic drop from $86.3\%$ AIA to $14.74\%$ MIA in Math) indicates they may offer incomplete or potentially misleading guidance when addressing student mistakes. These insights have profound implications for automated grading and assessment. The significant performance drop in MIA across all models raises concerns about the reliability of AI-driven systems in evaluating open-ended questions or complex problem-solving tasks. 

\textbf{Our work makes two key contributions to the field of LLMs and AI for education, with critical implications for developing more effective AI-powered tutoring systems.}
\textit{First}, we introduce MalAlgoQA in section~\ref{sec:dataset}, a novel dataset inspired by pedagogical distractor generation techniques, specifically designed to evaluate the counterfactual reasoning capabilities of LLMs. This approach mimics how educators create plausible but incorrect answer options, challenging LLMs to engage in the type of reasoning essential for understanding and addressing student misconceptions.
\textit{Second}, we propose the Malgorithm Identification task in section~\ref{sec:method}, along with two associated evaluation metrics: AIA and MIA. 
Our findings in sections~\ref{sec:exp_findings} and \ref{sec:analysis_ed} reveal a significant performance gap between these metrics across various LLMs, highlighting the challenge these models face in counterfactual reasoning.
We discuss the implications of our findings for AI-powered
educational tool in section~\ref{sec:implications}. 
By developing methods to evaluate LLMs' ability to identify and understand student misconceptions, our work paves the way for more effective AI tutoring systems that can accurately diagnose and address individual students' conceptual gaps \cite{nlet,sonkar-etal-2023-class,pedalign}.


\section{MalAlgoQA Dataset}
\label{sec:dataset}
We introduce MalAlgoQA, a real-world dataset consisting of multiple-choice questions in mathematics and reading comprehension for students in grade 3 through 11. 
The dataset is designed to evaluate the counterfactual reasoning abilities of LLMs through the task of analyzing the rationales and reasons behind incorrect answer choices, a process we refer to as ``Malgorithm identification.'' The MalAlgoQA dataset provides valuable resources to assess the counterfactual reasoning capabilities of LLMs through a pedagogical approach.

\subsection{Dataset Overview}

MalAlgoQA comprises two multiple-choice question sets: a mathematics and a reading comprehension question set. The mathematics question set contains 807 questions, while the reading comprehension question set has 290 questions. Each question is associated with four answer choices, as well as their corresponding rationales. Reading question set also provides a passage for each question. Table~\ref{tab:math_example} provides an example of math question, with more examples available in~\ref{app:example-1}.
Additionally, the dataset provides rich metadata including the grade level, content classification, and Depth of Knowledge (DOK) level for each question. 

\subsubsection{Grade Levels}
The questions in MalAlgoQA span across multiple grade levels, ranging from Grade 3 to Grade 11. As illustrated in Table \ref{tab:grade_distribution}, the distribution of questions for both mathematics and reading comprehension subsets is balanced across different grade levels.

\begin{table}[h]
\centering
\begin{tabular}{ccc}
\toprule
\textbf{Grade Level} & \textbf{Math} & \textbf{Reading}\\
\midrule
3  & 132 & 45 \\
4  & 120 & 47 \\
5  & 94  & 33 \\
6  & 140 & 41 \\
7  & 159 & 35 \\
8  & 128 & 42 \\
10 & -   & 47 \\
11 & 34  & -  \\
\bottomrule
\end{tabular}
\caption{Distribution of questions across grade levels.}
\label{tab:grade_distribution}
\end{table}

\subsubsection{Content Classifications}

The mathematics questions set are categorized into five content classifications: Number \& Operation, Algebra, Geometry \& Measurement, Data Analysis, and Data Analysis \& Probability. The reading comprehension questions are divided into two content classifications: Informational Text and Literature. Table \ref{tab:content_distribution} presents an overview of the distribution across content classifications for both question sets.

\begin{table}[h]
\centering
\begin{tabular}{cc}
\toprule
\multicolumn{2}{c}{\textbf{Math Content Classifications}} \\
\midrule
\textbf{Classification} & \textbf{Questions} \\
\midrule
Algebra & 315 \\
Number \& Operation & 312 \\
Geometry \& Measurement & 122 \\
Probability & 44 \\
Data Analysis & 14 \\
\bottomrule
\toprule
\multicolumn{2}{c}{\textbf{Reading Content Classifications}} \\
\midrule
\textbf{Classification} & \textbf{Questions} \\
\midrule
Informational Text & 190 \\
Literature & 100 \\
\bottomrule
\end{tabular}
\caption{Distribution of questions across content classifications.}
\label{tab:content_distribution}
\end{table}

\subsubsection{Depth of Knowledge (DOK) Levels}

Each question in MalAlgoQA is annotated with a DOK level, which indicates the cognitive complexity of the question. The DOK levels are derived from Norman L. Webb's taxonomy \cite{webb2002depth} and range from I to III, with higher levels representing increased cognitive complexity. Table \ref{tab:dok_distribution} shows the distribution of questions across DOK levels for both question sets.

\begin{table}[h!]
\centering
\begin{tabular}{ccc}
\toprule
\textbf{DOK Level} & \textbf{Math} & \textbf{Reading} \\
\midrule
I   & 349 & 37  \\
II  & 434 & 203 \\
III & 24  & 50  \\
\bottomrule
\end{tabular}
\caption{Distribution of questions across DOK levels.}
\label{tab:dok_distribution}
\end{table}

\subsection{Question and Rationale Characteristics}

The questions in MalAlgoQA vary in length and complexity. The average length of mathematics questions is 122 characters, while the average length of reading comprehension questions is 104 characters. Reading comprehension questions are accompanied by passages, which have an average length of 5783 characters. Each answer choice is associated with an rationale, which provide explanations for the correct and incorrect answer choices. The average length of answer rationales for mathematics questions is 227 characters, while the average length of answer rationales for reading comprehension questions is 1699 characters.

\section{Methodology}
\label{sec:method}
\begin{table*}[ht!]
\centering
\begin{minipage}[t]{0.9\textwidth}
\begin{lstlisting}[mathescape=true,basicstyle=\ttfamily\smalltonormalsize]
$\textbf{CoT Prompt for Malgorithm Identification}$: 

System: $\colorbox{lightyellow}{Provide step-by-step reasoning}$ to determine the answer rationale that corresponds to the given choice for the question. Then, provide your answer in the specified JSON format.

Question: {question}
Choice: {choice}
Answer Rationales: {formatted_rationales}

Given the question and answer rationales, determine which answer rationale corresponds to the given choice for the question. 
$\colorbox{lightyellow}{Provide step-by-step reasoning. Show the steps.}$
And then provide your answer in the following JSON format:
{{
  "Correct Choice": "[A/B/C/D]"
}}
\end{lstlisting}
\end{minipage}
\vspace{-2mm}
\caption{Prompt for the Malgorithm Identification experiment, where the LLM identifies the relationship between answer choices and rationales in a multiple-choice question-answering setting. The task involves identifying the ``malgorithm'', a rationale that represents flawed reasoning steps which lead to an incorrect answer choice. The prompt provides the question, an answer choice (either correct or incorrect), and all rationales, asking the LLM to identify the rationale corresponding to the given choice. The highlighted sentence instructs the model to provide step-by-step reasoning, an instruction added for the Chain-of-Thought prompting strategy.}
\label{tab:cot_prompts}
\end{table*}

In this section, we present the task formulation for Malgorithm Identification.

\subsection{Definitions and Task Formulation}
Let $Q = \{q_1, q_2, ..., q_n\}$ be a set of multiple-choice questions, where each question $q_i$ has a set of answer choices $C_i = \{c_{i1}, c_{i2}, c_{i3}, c_{i4}\}$ and a set of rationales $R_i = \{r_{i1}, r_{i2}, r_{i3}, r_{i4}\}$ corresponding to each answer choice. The rationale can be thought of as a chain of reasoning steps, where each step represents a part of the thought process leading to the selection of an answer choice.

\myparagraph{Malgorithm}
We introduce the term ``malgorithm'' to describe a rationale underlying an incorrect answer choice. A malgorithm is composed of a series of reasoning steps, some of which are correct, while others contain errors. We refer to these erroneous steps as ``mal-rules''.
The term mal-rules is derived from the cognitive error literature \citep{brown1980repair,vanlehn1990mind,payne1990algebra}, which refers to the flawed or incorrect rules that students may apply when solving problems, resulting in systematic errors. In our context, a malgorithm is a reasoning chain that consists of both correct rules and mal-rules, with the presence of mal-rules ultimately leading to the selection of an incorrect answer choice.


\myparagraph{Malgorithm Identification Task}
Given a question $q_i$, an answer choice $c_{ij}$ (either correct or incorrect), and the set of rationales $R_i$, the LLM must identify the rationale $r_{ik}$ that corresponds to the given answer choice $c_{ij}$. If the given answer choice is incorrect, the corresponding rationale is a malgorithm containing a combination of correct reasoning steps and one or more mal-rules.

\subsection{Evaluation Metrics}
\label{sec:metrics}

To gain a comprehensive understanding of the LLMs performance in the Malgorithm Identification task, we use two key metrics:

1. \textbf{Algorithm Identification Accuracy (AIA)}: This metric measures the model's ability to identify the rationale or algorithm that corresponds to the correct answer choice. Essentially, AIA evaluates how well the model recognizes and follows a chain of reasoning steps that lead to the correct answer.

2. \textbf{Malgorithm Identification Accuracy (MIA)}: This metric measures the model's ability to identify the flawed reasoning or malgorithm that corresponds to an incorrect answer choice. Specifically, MIA evaluates the model's capacity to identify faulty reasoning chains that contain a combination of correct steps and one or more mal-rules, which lead to the wrong answer.



\section{Experiments and Findings}
\label{sec:exp_findings}
\begin{table*}[t!]
\centering
\begin{tabular}{ccccccc}
\toprule
\textbf{LLM} & \textbf{Subject} & \textbf{Num Shots} & \textbf{Prompting} & \textbf{MCQ} & \textbf{AIA} & \textbf{MIA} \\
\midrule
\multirow{2}{*}{GPT 4o}  & \multirow{2}{*}{Reading} & 0 & Simple & 95.86 & 98.23 & 82.80 \\
  &  & 0 & CoT & 95.52 & 97.88 & 68.63 \\
\midrule
\multirow{2}{*}{LLaMA3-70B}  & \multirow{2}{*}{Reading} & 0 & Simple & 89.45 & 95.72 & 60.00 \\
  &  & 0 & CoT & 89.27 & 94.60 & 58.40 \\
\midrule
\multirow{2}{*}{GPT-3.5}  & \multirow{2}{*}{Reading} & 0 & Simple & 84.83 & 95.05 & 13.98 \\
  &  & 0 & CoT & 81.66 & 85.82 & 8.61 \\
\midrule
\multirow{2}{*}{LLaMA3-8B}  & \multirow{2}{*}{Reading} & 0 & Simple & 73.96 & 89.62 & 19.89 \\
  &  & 0 & CoT & 78.05 & 89.52 & 15.47 \\
\bottomrule
\toprule
\multirow{4}{*}{GPT 4o}  & \multirow{4}{*}{Math} & 0 & Simple & 90.11 & 92.06 & 62.82 \\
 &  & 5 & Simple &  & 90.61 & 64.99 \\
  &  & 0 & CoT & 97.61 & 95.65 & 66.10 \\
  &  & 5 & CoT &  & 95.49 & 65.10 \\
\midrule
\multirow{4}{*}{LLaMA3-70B}  & \multirow{4}{*}{Math} & 0 & Simple & 73.83 & 94.31 & 50.22 \\
  &  & 5 & Simple &  & 93.65 & 48.31 \\
  &  & 0 & CoT & 90.72 & 95.47 & 55.32 \\
  &  & 5 & CoT &  & 95.89 & 50.6 \\
\midrule
\multirow{4}{*}{GPT 3.5}  & \multirow{4}{*}{Math} & 0 & Simple & 52.85 & 77.08 & 35.8 \\
&  & 5 & Simple &  & 76.85 & 34.61 \\
  &  & 0 & CoT & 82.53 & 86.28 & 14.74 \\
  &  & 5 & CoT &  & 81.87 & 17.06 \\
\midrule
\multirow{4}{*}{LLaMA3-8B}  & \multirow{4}{*}{Math} & 0 & Simple & 45.79 & 75.29 & 34.81 \\
 &  & 5 & Simple &  & 80.32 & 38.41 \\
 &  & 0 & CoT & 70.68 & 83.07 & 27.05 \\
 &  & 5 & CoT &  & 83.05 & 31.39 \\
\bottomrule
\end{tabular}
\caption{Performance results of LLMs on the Malgorithm Identification task for Reading and Math subjects. Results are organized by prompting type (Simple/CoT) and number of shots (0/5). Metrics include Multiple-Choice Question (MCQ) accuracy, Algorithm Identification Accuracy (AIA), and Malgorithm Identification Accuracy (MIA). AIA measures correct rationale identification, while MIA measures flawed reasoning identification. MIA scores are generally significantly lower than AIA, highlighting the challenge of counterfactual reasoning.}
\label{tab:results}
\end{table*}

\begin{table*}[t]
\begin{minipage}{.49\linewidth}
\begin{tabular}{cccc}
\toprule
\textbf{Subject} & \textbf{Grade} & \textbf{GPT 4o} & \textbf{LLaMA3-70B}\\
\midrule
\multirow{7}{*}{\textbf{Math}} & 3  & \underline{67.00} & 44.26 \\
& 4  & \underline{65.93} & \underline{54.13} \\
& 5  & \underline{66.52}  & 46.97 \\
& 6  & \underline{67.80} & \underline{58.84} \\
& 7  & 64.50 & \underline{57.49} \\
& 8  & 62.96 & 42.44 \\
& 11 & 53.68  & 41.03  \\
\midrule
& average & 65.10 & 50.60\\
\bottomrule
\end{tabular}
\end{minipage}
\begin{minipage}{.49\linewidth}
\begin{tabular}{cccc}
\toprule
\textbf{Subject} & \textbf{Grade} & \textbf{GPT 4o} & \textbf{LLaMA3-70B}\\
\midrule
\multirow{7}{*}{\textbf{Reading}} & 3  & \underline{73.81} & \underline{65.18} \\
& 4  & \underline{68.89} &  \underline{60.71} \\
& 5  & \underline{69.47}  & 45.68 \\
& 6  & 58.62 & 50.00 \\
& 7  & 63.46 & \underline{80.22} \\
& 8  & \underline{80.67} & \underline{62.00} \\
& 10 & 64.96  &  46.83 \\
\midrule
& average & 68.63 & 58.40 \\
\bottomrule
\end{tabular}
\end{minipage}
\caption{Grade-level performance distribution (MIA) for GPT-4o and LLaMA3-70B models using Chain-of-Thought prompting (few-shot for Math, zero-shot for Reading). Grades with above-average performance are \underline{\textbf{underlined}}.}
\label{tab:grade_score}
\end{table*}

\subsection{Experimental Setup}
We conducted experiments on the Malgorithm Identification task with four state-of-the-art LLMs: GPT-4o, GPT-3.5, LLaMA3-70B and LLaMA3-8B.\footnote{The specific models we use are GPT-4o: gpt-4o-2024-05-13, GPT-3.5: gpt-3.5-turbo-0125, LLaMA3-70B: Meta-LLaMA3-70B-Instruct, LLaMA3-8B: Meta-LLaMA3-8B-Instruct} To evaluate the LLMs performance, we use AIA and MIA metrics (described in section~\ref{sec:metrics}).

For each model, we evaluated with four experimental settings: (1) Simple prompting, where the LLMs are prompted to provide only the final answer; (2) Chain-of-Thought (CoT) prompting \cite{cot,letsthinkstepbystep}, where the LLMs are prompted to provide step-by-step reasoning and show the intermediate steps before giving the final answer, as shown in Table~\ref{tab:cot_prompts};  (3) Zero-shot learning, where the LLMs are provided without any additional examples; (4) Few-shot learning, where the LLMs are provided with a small number of examples demonstrating the desired reasoning process, in addition to the standard prompts. Due to context length constraints, we conducted few-shot experiments only on the Math dataset.

\subsection{Main Findings}
\myparagraph{Wide Performance Gap Between MIA and AIA}
The most striking observation from Table~\ref{tab:results} is the substantial performance gap between Malgorithm Identification Accuracy (MIA) and Algorithm Identification Accuracy (AIA) across all models. For GPT-4o on the Math dataset with zero-shot CoT prompting, we see an AIA of $95.65\%$ compared to an MIA of only $66.10\%$ - a drop of nearly $30$ percentage points. This pattern is consistent across models and prompting strategies, with the gap widening for smaller models. For instance, GPT-3.5 shows an even more dramatic difference: $86.28\%$ AIA vs $ 14.74\%$ MIA.

This disparity suggests that while LLMs have become proficient at identifying correct reasoning paths, they struggle significantly when asked to recognize and articulate flawed reasoning. This challenge goes beyond simple counterfactual reasoning and points to a fundamental limitation in how these models process and evaluate logical structures. The ability to identify correct reasoning does not necessarily translate to the ability to pinpoint errors in incorrect reasoning, indicating a potential blind spot in the training or architecture of current LLMs.

\myparagraph{Impact of Model Size on MIA Performance}
The data reveals a clear correlation between model size and MIA performance. For the Math dataset with zero-shot CoT prompting, we see a progression from LLaMA3-8B ($27.05\%$ MIA) to LLaMA3-70B ($55.32\%$ MIA) to GPT-4o ($66.10\%$ MIA). This trend suggests that the ability to identify malgorithms is an emergent property that significantly improves with increased model capacity \cite{wei2022emergent}.
However, it's important to note that this improvement with model size is much more pronounced for MIA than for AIA. While GPT-4o outperforms LLaMA3-8B by $39.05$ percentage points in MIA, the difference in AIA is only $12.58$ percentage points ($95.65\%$ vs $83.07\%$). This disproportionate scaling indicates that malgorithm identification may require more complex reasoning capabilities that are only unlocked at larger model scales.

\begin{figure*}[t!]
\begin{minipage}[t]{0.50\linewidth}
\begin{center}
\includegraphics[width=1\linewidth]{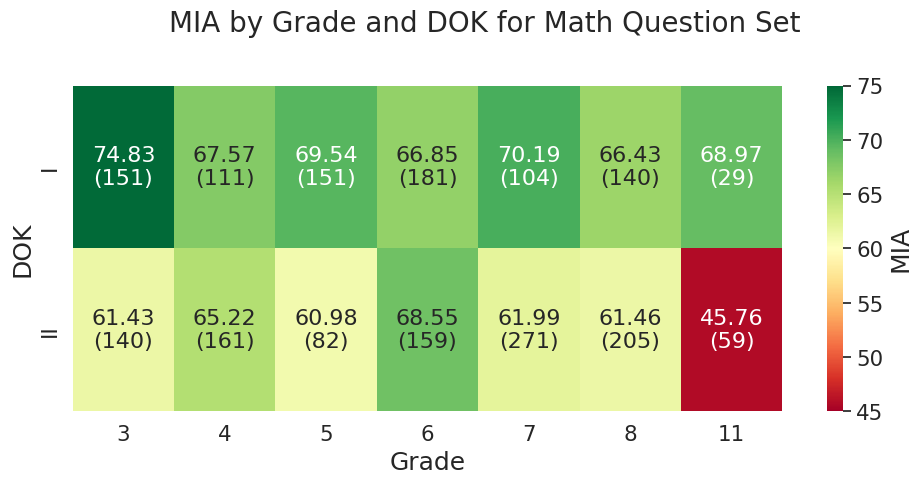} 
\end{center} 
\end{minipage}
\hfill
\begin{minipage}[t]{0.50\linewidth}
\begin{center}
\includegraphics[width=1\linewidth]{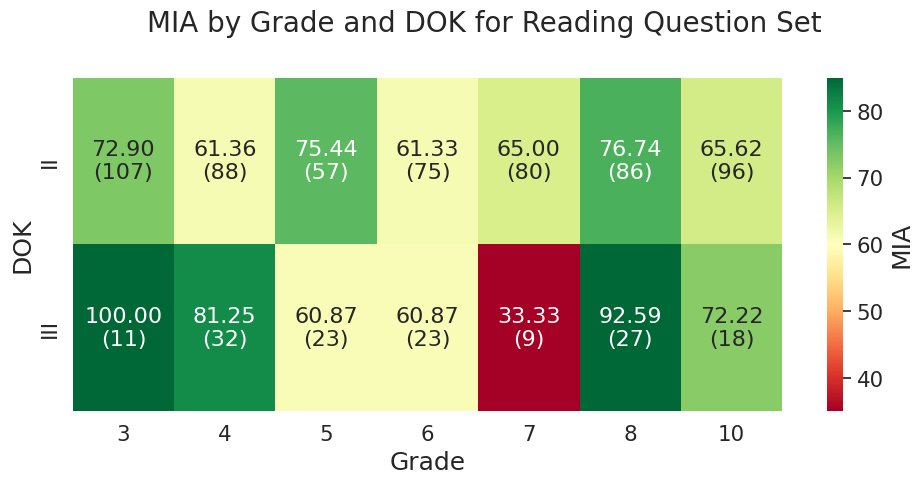} 
\end{center}
\end{minipage}
\vspace{-5mm}
\caption{An illustration of MIA performance across different Grades and DOK levels for Math and Reading question sets. The top number represents MIA performance and the bottom number represents question counts. DOK I and II are selected for Math and DOK II and III for Reading. (best viewed in colors).}
\label{dok}
\end{figure*}

\myparagraph{Ineffectiveness of Few-Shot Learning}
The data clearly demonstrates that few-shot learning does not significantly improve MIA performance across models. For GPT-4o on the Math dataset, the MIA score with CoT prompting actually decreases slightly from $66.10\%$ (zero-shot) to $65.10\%$ (5-shot). LLaMA3-70B shows a similar trend, with MIA dropping from $55.32\%$ to $50.6\%$.
Even in cases where there is a marginal improvement, the gains are negligible. For instance, LLaMA3-8B's MIA increases from $27.05\%$ to $31.39\%$ with 5-shot CoT prompting - a difference of only $4.34$ percentage points. GPT-3.5 shows a similarly small improvement from $14.74\%$ to $17.06\%$.

These results suggest that providing examples does not substantially enhance an LLM's ability to identify malgorithms. This finding is particularly noteworthy given that few-shot learning often improves performance on other NLP tasks. The ineffectiveness of this approach for malgorithm identification underscores the unique challenges posed by this task and suggests that more sophisticated techniques may be needed to improve LLM performance in this area.

\subsection{Understanding CoT Prompting for MIA}

Our analysis of Chain-of-Thought (CoT) prompting reveals an unexpected ineffectiveness in improving Malgorithm Identification Accuracy (MIA), and in some cases, even leads to decreased performance compared to simple prompting.

\textbf{Inconsistent impact across subjects:} For GPT-4o, CoT prompting shows negligible difference in Math MIA ($65.1\%$ vs $65.0\%$ for simple prompting), but significantly underperforms in Reading MIA ($68.6\%$ vs $82.8\%$ for simple prompting).

\textbf{Consistent underperformance of LLaMA-70B:} LLaMA-70B demonstrates lower MIA prompting in both Reading ($58.4\%$ vs $60.0\%$) and Math ($48.3\%$ vs $50.6\%$) with CoT compared to simple prompting.

\textbf{Minimal impact on AIA:} CoT prompting generally performs similarly to simple prompting for Algorithm Identification Accuracy (AIA) across models and subjects, indicating its ineffectiveness is specific to malgorithm identification.

\begin{figure*}[th!]
\centering
\includegraphics[width=0.7\linewidth]{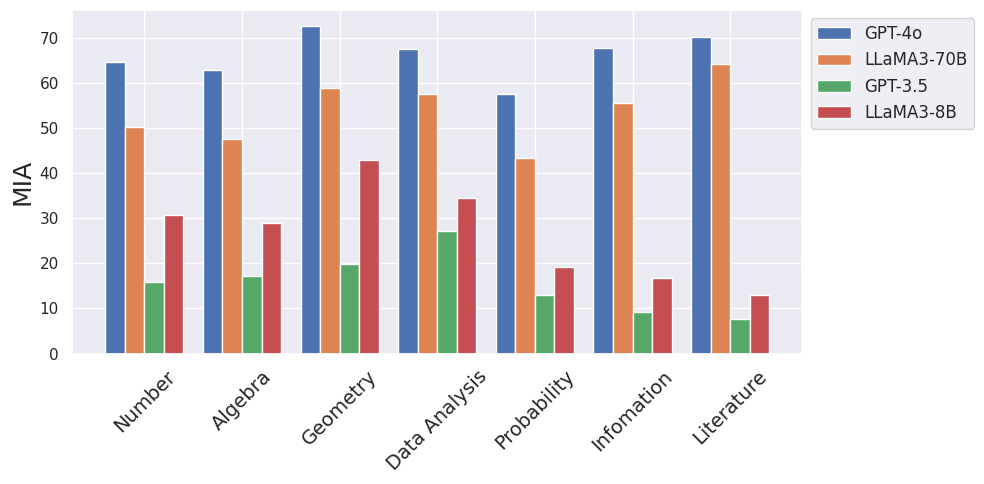} 
\hfill
\caption{An illustration of MIA performance across five math content classes (Number \& Operation, Algebra, Geometry \& Measurement, Data Analysis, Probability) and two Reading content classes (Informational, Literature).}
\label{figcontent}
\end{figure*}

These results challenge the assumption that explicit step-by-step reasoning always improves performance on complex tasks. The data suggests that CoT prompting, while effective for many NLP tasks, may actually hinder the model's ability to identify flawed reasoning paths in the context of malgorithm identification.

A plausible explanation for this counterintuitive finding lies in the nature of the training data typically used for LLMs. These models are predominantly exposed to examples of correct reasoning during pre-training and fine-tuning. Consequently, when CoT prompting guides the model through a structured thought process, it likely leverages these learned patterns of correct reasoning. This may inadvertently anchor the model to correct logical pathways, making it more challenging to deviate from these patterns and identify flawed reasoning.
In essence, CoT prompting might be reinforcing the model's bias towards correct reasoning, which is counterproductive when the task specifically requires identifying incorrect logical steps. 



\section{Analysis: Educational Dimensions}
\label{sec:analysis_ed}

\myparagraph{Performance across Grade Levels} 
Table~\ref{tab:grade_score} analyzes the MIA performance of GPT-4o and LLaMA3-70B models across different grade levels in math and reading datasets. Both models consistently show a decline in performance with increasing grade levels, with LLaMA3-70B exhibiting some unexpected variability, particularly excelling in Grade 8 of the reading dataset. These results highlight the increasing challenges in tackling malgorithm identification task as the questions becomes more complex. 

\myparagraph{Performance across DOK}
Figure~\ref{dok} presents a comprehensive overview of performance across DOK levels for various Grade using MIA score from GPT 4o with CoT prompting strategy. The pattern indicates that as the DOK level and Grade increases, MIA performance tends to decline, especially highlighted by the comparison between the top-left cell (dark green) and bottom-right cell (dark red). Similarly, in the Reading dataset, performance at DOK Level III drops to 33.33 for Grade 7. This trend further emphasizes the inherent difficulty in evaluating the malgorithms of questions that require deeper understanding and analytical thinking.

\myparagraph{Performance across Content Classifications}
Figure~\ref{figcontent} illustrates the performance of LLMs in detecting malgorithms across five math contents and two reading content classifications. 
For the math dataset, Geometry consistently yields the highest performance across all models, suggesting that even for malgorithm identification, structural content like Geometry is easier for LLMs to handle as compared to numerical content. On the other hand, Probability appears to be more challenging content, reflecting the inherent difficulty of LLMs in probabilistic reasoning. 
For the reading dataset, GPT-4o and LLaMA3-70B excel at Informational Text, while smaller models like GPT-3.5 and LLaMA3-8B perform better with Literature. This observation suggests that larger models are better at processing structured, factual content, whereas smaller models are more efficient with narrative, context-rich material.

\section{Implications for AI in Education}
\label{sec:implications}
The findings from our study on malgorithm identification have significant implications for the development and application of AI in educational contexts:

\textbf{Limitations in Understanding Student Misconceptions:} The substantial performance gap between Algorithm Identification Accuracy (AIA) and Malgorithm Identification Accuracy (MIA) suggests that current LLMs are far more adept at recognizing correct reasoning than identifying flawed logic. For GPT-4o, we observed a drop from $95.65\%$ AIA to $66.10\%$ MIA on the Math dataset. This disparity indicates that AI tutoring systems based on these models may excel at confirming correct answers but struggle to diagnose and address student misconceptions effectively. This limitation could lead to missed opportunities for targeted intervention and personalized learning.

\textbf{Rethinking Training Approaches for Error Recognition}
The ineffectiveness of CoT prompting in improving MIA performance reveals a fundamental challenge in training LLMs to recognize and analyze errors. Rather than simply diversifying training data, this finding suggests a need for novel training paradigms that specifically target the skill of error identification \cite{sonkar2024regressive}. Future AIED systems may require specialized fine-tuning processes or architectural modifications that enable models to simultaneously reason about correct and incorrect logical pathways. This could involve developing new pre-training tasks focused on error detection, or creating model architectures that maintain separate representations for correct and flawed reasoning patterns. Such innovations could lead to AI tutors that are equally adept at reinforcing correct understanding and diagnosing misconceptions, significantly enhancing their effectiveness in personalized education.

\textbf{Rethinking Feedback Mechanisms:} The challenges in malgorithm identification suggest that current AI models might provide incomplete or misleading feedback to students. AIED systems need to be designed with an awareness of these limitations, potentially incorporating human oversight or alternative assessment methods to ensure that student misconceptions are accurately identified and addressed.

\textbf{Implications for Automated Grading and Assessment:} The performance discrepancy between AIA and MIA also raises concerns about the reliability of AI-driven automated grading systems, particularly for open-ended questions or problems requiring complex reasoning. Educational institutions considering the implementation of such systems should be aware of these limitations and consider alternate \cite{marking,alag} or hybrid approaches with human review.

\textbf{Adaptive Difficulty Handling:} The observed decline in MIA performance across increasing grade levels and DOK levels highlights the need for AIED systems that can adapt to varying levels of question complexity. Future AI tutors should be designed with mechanisms to maintain consistent performance across different difficulty levels, ensuring effective support for students as they progress through more challenging material.

The difficulty LLMs face in identifying flawed reasoning points to a broader challenge in counterfactual reasoning. For AIED applications, this suggests a need to develop models with enhanced capabilities in exploring alternative scenarios and reasoning paths. Such improvements could lead to more sophisticated tutoring systems that can not only identify errors but also explain why certain approaches are incorrect and guide students towards correct understanding. By addressing these implications, AIED researchers can work towards not only improving the accuracy of student assessment but also enhancing the ability of AI systems to provide targeted and personalized support.

\section{Related Work}
Counterfactual reasoning, a critical cognitive process for understanding causality, has gained significant attention in machine learning (ML) and NLP research. Pearl's seminal work on causal inference \cite{pearl2009causality} has been widely applied to address challenges in fairness \cite{kusner2017counterfactual}, interpretability \cite{wachter2017counterfactual}, and robustness \cite{bareinboim2016causal} in ML. In NLP, counterfactual reasoning is important for tasks requiring comprehension of hypothetical scenarios. However, most existing NLP models primarily focus on predictive tasks rather than hypothetical reasoning \cite{bommasani2021opportunities, bender2021dangers}. Recent work has begun to address this gap developing specialized datasets and models for counterfactual reasoning \cite{ qin2019counterfactual}. Several datasets have been created to evaluate different aspects of reasoning, including counterfactual reasoning such as COPA dataset \cite{roemmele2011choice}, Counterfactual Story Rewriting task \cite{qin2020back}, and CQA dataset \cite{talmor2019commonsenseqa}. These efforts are complemented by development of new modeling approaches, such as causal LMs \cite{feder2021causalm}, counterfactual data augmentation \cite{kaushik2020learning}, and generating counterfactual explanations \cite{wu2021polyjuice, ross2021explaining}. Despite these advancements, challenges remain such as handling multi-step causal chains and generalizing across diverse domains \cite{yao2021counterfactual, keith2020text, feder2022causal}.

\section{Conclusion and Future Work}
This paper introduces MalAlgoQA, a novel dataset designed to evaluate the counterfactual reasoning capabilities of LLMs through a pedagogical lens. By proposing the Malgorithm Identification task and its associated metrics, we provide a new framework for assessing LLMs' ability to recognize and analyze flawed reasoning patterns. Our findings reveal significant challenges in counterfactual reasoning for state-of-the-art LLMs. These results highlight the need for further research into enhancing LLMs' counterfactual reasoning abilities, particularly for educational applications where understanding student misconceptions is crucial. Future work will focus on expanding the dataset, exploring novel training methodologies, and investigating architectural modifications to develop more cognitively-aligned language models capable of robust counterfactual reasoning.


\section{Limitations}
While our research sheds light on the counterfactual reasoning capabilities of LLMs, we acknowledge the inherent complexity and intricacy of the cognitive process may not be fully captured by our current evaluation framework. The MalAlgoQA dataset and the Malgorithm Identification task, although designed with careful consideration, are initial steps toward understanding the intricate aspect of cognition. The chain-of-thought (CoT) prompting technique is not a panacea and its effectiveness vary depending on the specific context and task. As research in this field progresses, we anticipate further refinements and enhancements to our methodology, contributing to a more comprehensive understanding of counterfactual reasoning in LLMs.

\section{Ethics and Risk}
The use of LLMs in classrooms, for instance, could profoundly influence the learning experiences of students. While LLMs can provide personalized learning support and instant feedback, their inability to accurately identify and address student misconceptions, as highlighted by our study, could potentially reinforce incorrect understanding or reasoning. However, the transparency of LLMs' decision-making process is a crucial ethical consideration. If LLMs are to be trusted educational aids, it is critical for students, teachers, and parents to understand the reasoning behind the LLMs' responses. Our work with the MalAlgoQA dataset is a significant step towards understanding and addressing these ethical considerations. By examining the counterfactual causal reasoning abilities of LLMs, we can assess their readiness for use in educational settings and contribute to the ongoing discourse surrounding the ethical deployment of AI in education. However, we acknowledge that this is a complex, multifaceted issue that requires continuous exploration, vigilance, and open dialogue among researchers, educators, policymakers, and society at large.

\section*{Acknowledgments}
This work was supported by NSF grant 1842378, ONR grant N0014-20-1-2534, AFOSR grant FA9550-22-1-0060, a Vannevar Bush Faculty Fellowship, and ONR grant N00014-18-1-2047.

\bibliography{custom}

\begin{thebibliography}{36}
\providecommand{\natexlab}[1]{#1}

\bibitem[{Bareinboim and Pearl(2016)}]{bareinboim2016causal}
Elias Bareinboim and Judea Pearl. 2016.
\newblock Causal inference and the data-fusion problem.
\newblock \emph{Proceedings of the National Academy of Sciences}, 113(27):7345--7352.

\bibitem[{Bender et~al.(2021)Bender, Gebru, McMillan-Major, and Shmitchell}]{bender2021dangers}
Emily~M Bender, Timnit Gebru, Angelina McMillan-Major, and Margaret Shmitchell. 2021.
\newblock On the dangers of stochastic parrots: Can language models be too big?
\newblock \emph{Proceedings of the 2021 ACM Conference on Fairness, Accountability, and Transparency}, pages 610--623.

\bibitem[{Bommasani et~al.(2021)Bommasani, Hudson, Adeli, Altman, Arora, von Arx, Bernstein, Bohg, Bosselut, Brunskill et~al.}]{bommasani2021opportunities}
Rishi Bommasani, Drew~A Hudson, Ehsan Adeli, Russ Altman, Simran Arora, Sydney von Arx, Michael~S Bernstein, Jeannette Bohg, Antoine Bosselut, Emma Brunskill, et~al. 2021.
\newblock On the opportunities and risks of foundation models.
\newblock \emph{arXiv preprint arXiv:2108.07258}.

\bibitem[{Brown and VanLehn(1980)}]{brown1980repair}
John~Seely Brown and Kurt VanLehn. 1980.
\newblock Repair theory: A generative theory of bugs in procedural skills.
\newblock \emph{Cognitive science}, 4(4):379--426.

\bibitem[{Dubey et~al.(2024)Dubey, Jauhri, Pandey, Kadian, Al-Dahle, Letman, Mathur, Schelten, Yang, Fan et~al.}]{llama3modelcard}
Abhimanyu Dubey, Abhinav Jauhri, Abhinav Pandey, Abhishek Kadian, Ahmad Al-Dahle, Aiesha Letman, Akhil Mathur, Alan Schelten, Amy Yang, Angela Fan, et~al. 2024.
\newblock The llama 3 herd of models.
\newblock \emph{arXiv preprint arXiv:2407.21783}.

\bibitem[{Feder et~al.(2021)Feder, Oved, Shalit, and Reichart}]{feder2021causalm}
Amir Feder, Nadav Oved, Uri Shalit, and Roi Reichart. 2021.
\newblock Causalm: Causal model explanation through counterfactual language models.
\newblock \emph{Computational Linguistics}, 47(2):333--386.

\bibitem[{Feder et~al.(2022)Feder, Oved, Shalit, and Reichart}]{feder2022causal}
Amir Feder, Nadav Oved, Uri Shalit, and Roi Reichart. 2022.
\newblock Causal inference in natural language processing: Estimation, prediction, interpretation and beyond.
\newblock \emph{Transactions of the Association for Computational Linguistics}, 10:1045--1063.

\bibitem[{Kaushik et~al.(2020)Kaushik, Hovy, and Lipton}]{kaushik2020learning}
Divyansh Kaushik, Eduard Hovy, and Zachary~C Lipton. 2020.
\newblock Learning the difference that makes a difference with counterfactually-augmented data.
\newblock In \emph{International Conference on Learning Representations}.

\bibitem[{Keith et~al.(2020)Keith, Jensen, and O'Connor}]{keith2020text}
Katherine~A Keith, David Jensen, and Brendan O'Connor. 2020.
\newblock Text and causal inference: A review of using text to remove confounding from causal estimates.
\newblock \emph{arXiv preprint arXiv:2005.00649}.

\bibitem[{Kojima et~al.(2022)Kojima, Gu, Reid, Matsuo, and Iwasawa}]{letsthinkstepbystep}
Takeshi Kojima, Shixiang~Shane Gu, Machel Reid, Yutaka Matsuo, and Yusuke Iwasawa. 2022.
\newblock Large language models are zero-shot reasoners.
\newblock \emph{Advances in neural information processing systems}, 35:22199--22213.

\bibitem[{Kusner et~al.(2017)Kusner, Loftus, Russell, and Silva}]{kusner2017counterfactual}
Matt~J Kusner, Joshua Loftus, Chris Russell, and Ricardo Silva. 2017.
\newblock Counterfactual fairness.
\newblock In \emph{Advances in Neural Information Processing Systems}, pages 4066--4076.

\bibitem[{Liu et~al.(2023)Liu, Sonkar, Wang, Woodhead, and Baraniuk}]{nlet}
Naiming Liu, Shashank Sonkar, Zichao Wang, Simon Woodhead, and Richard~G Baraniuk. 2023.
\newblock Novice learner and expert tutor: Evaluating math reasoning abilities of large language models with misconceptions.
\newblock \emph{arXiv preprint arXiv:2310.02439}.

\bibitem[{OpenAI(2023)}]{openai2023gpt4}
OpenAI. 2023.
\newblock \href {https://arxiv.org/abs/2303.08774} {Gpt-4 technical report}.
\newblock \emph{Preprint}, arXiv:2303.08774.

\bibitem[{Payne and Squibb(1990)}]{payne1990algebra}
Stephen~J Payne and Helen~R Squibb. 1990.
\newblock Algebra mal-rules and cognitive accounts of error.
\newblock \emph{Cognitive science}, 14(3):445--481.

\bibitem[{Pearl(2009)}]{pearl2009causality}
Judea Pearl. 2009.
\newblock \emph{Causality}.
\newblock Cambridge University Press.

\bibitem[{Pearl(2019)}]{pearl2019seven}
Judea Pearl. 2019.
\newblock The seven tools of causal inference, with reflections on machine learning.
\newblock \emph{Communications of the ACM}, 62(3):54--60.

\bibitem[{Qin et~al.(2019)Qin, Bosselut, Holtzman, Bhagavatula, Clark, and Choi}]{qin2019counterfactual}
Lianhui Qin, Antoine Bosselut, Ari Holtzman, Chandra Bhagavatula, Elizabeth Clark, and Yejin Choi. 2019.
\newblock Counterfactual story reasoning and generation.
\newblock In \emph{Proceedings of the 2019 Conference on Empirical Methods in Natural Language Processing and the 9th International Joint Conference on Natural Language Processing}, pages 5043--5053.

\bibitem[{Qin et~al.(2020)Qin, Bosselut, Holtzman, Bhagavatula, Clark, and Choi}]{qin2020back}
Lianhui Qin, Antoine Bosselut, Ari Holtzman, Chandra Bhagavatula, Elizabeth Clark, and Yejin Choi. 2020.
\newblock Back to the future: Unsupervised backprop-based decoding for counterfactual and abductive commonsense reasoning.
\newblock In \emph{Proceedings of the 2020 Conference on Empirical Methods in Natural Language Processing}, pages 794--805.

\bibitem[{Rafetseder et~al.(2021)Rafetseder, Cristi-Vargas, and Perner}]{rafetseder2021counterfactual}
Eva Rafetseder, Renato Cristi-Vargas, and Josef Perner. 2021.
\newblock Counterfactual reasoning in children.
\newblock \emph{Journal of Experimental Child Psychology}, 204:105054.

\bibitem[{Roemmele et~al.(2011)Roemmele, Bejan, and Gordon}]{roemmele2011choice}
Melissa Roemmele, Cosmin~Adrian Bejan, and Andrew~S Gordon. 2011.
\newblock Choice of plausible alternatives: An evaluation of commonsense causal reasoning.
\newblock In \emph{2011 AAAI Spring Symposium Series}.

\bibitem[{Ross et~al.(2021)Ross, Marasovi{\'c}, and Peters}]{ross2021explaining}
Alexis Ross, Ana Marasovi{\'c}, and Matthew~E Peters. 2021.
\newblock Explaining nlp models via minimal contrastive editing (mice).
\newblock In \emph{Findings of the Association for Computational Linguistics: ACL-IJCNLP 2021}, pages 3840--3852.

\bibitem[{Schulman et~al.(2022)Schulman, Zoph, Kim, Hilton, Menick, Weng, Uribe, Fedus, Metz, Pokorny et~al.}]{schulman2022chatgpt}
J~Schulman, B~Zoph, C~Kim, J~Hilton, J~Menick, J~Weng, JFC Uribe, L~Fedus, L~Metz, M~Pokorny, et~al. 2022.
\newblock Chatgpt: Optimizing language models for dialogue.

\bibitem[{Sonkar and Baraniuk(2023)}]{dupe}
Shashank Sonkar and Richard~G Baraniuk. 2023.
\newblock {Deduction under Perturbed Evidence: Probing Student Simulation (Knowledge Tracing) Capabilities of Large Language Models.}
\newblock In \emph{LLM@ AIED}, pages 26--33.

\bibitem[{Sonkar et~al.(2024{\natexlab{a}})Sonkar, Liu, and Baraniuk}]{sonkar2024regressive}
Shashank Sonkar, Naiming Liu, and Richard~G Baraniuk. 2024{\natexlab{a}}.
\newblock Regressive side effects of training language models to mimic student misconceptions.
\newblock \emph{arXiv preprint arXiv:2404.15156}.

\bibitem[{Sonkar et~al.(2023)Sonkar, Liu, Mallick, and Baraniuk}]{sonkar-etal-2023-class}
Shashank Sonkar, Naiming Liu, Debshila Mallick, and Richard Baraniuk. 2023.
\newblock \href {https://doi.org/10.18653/v1/2023.findings-emnlp.130} {{{CLASS}: A Design Framework for Building Intelligent Tutoring Systems Based on Learning Science principles}}.
\newblock In \emph{Findings of the Association for Computational Linguistics: EMNLP 2023}, pages 1941--1961, Singapore. Association for Computational Linguistics.

\bibitem[{Sonkar et~al.(2024{\natexlab{b}})Sonkar, Liu, Mallick, and Baraniuk}]{marking}
Shashank Sonkar, Naiming Liu, Debshila~B. Mallick, and Richard~G. Baraniuk. 2024{\natexlab{b}}.
\newblock {{Marking: Visual Grading with Highlighting Errors and Annotating Missing Bits}}.
\newblock In \emph{{{Artificial Intelligence in Education}}}, pages 309--323, Cham. Springer Nature Switzerland.

\bibitem[{Sonkar et~al.(2024{\natexlab{c}})Sonkar, Ni, Chaudhary, and Baraniuk}]{pedalign}
Shashank Sonkar, Kangqi Ni, Sapana Chaudhary, and Richard~G. Baraniuk. 2024{\natexlab{c}}.
\newblock \href {https://arxiv.org/abs/2402.05000} {{Pedagogical Alignment of Large Language Models}}.
\newblock \emph{Preprint}, arXiv:2402.05000.

\bibitem[{Sonkar et~al.(2024{\natexlab{d}})Sonkar, Ni, Tran~Lu, Kincaid, Hutchinson, and Baraniuk}]{alag}
Shashank Sonkar, Kangqi Ni, Lesa Tran~Lu, Kristi Kincaid, John~S. Hutchinson, and Richard~G. Baraniuk. 2024{\natexlab{d}}.
\newblock {{Automated Long Answer Grading with RiceChem Dataset}}.
\newblock In \emph{Artificial Intelligence in Education}, pages 163--176, Cham. Springer Nature Switzerland.

\bibitem[{Talmor et~al.(2019)Talmor, Herzig, Lourie, and Berant}]{talmor2019commonsenseqa}
Alon Talmor, Jonathan Herzig, Nicholas Lourie, and Jonathan Berant. 2019.
\newblock Commonsenseqa: A question answering challenge targeting commonsense knowledge.
\newblock In \emph{Proceedings of the 2019 Conference of the North American Chapter of the Association for Computational Linguistics: Human Language Technologies}, pages 4149--4158.

\bibitem[{VanLehn(1990)}]{vanlehn1990mind}
Kurt VanLehn. 1990.
\newblock Mind bugs: The origins of procedural misconceptions.
\newblock \emph{MIT press}.

\bibitem[{Wachter et~al.(2018)Wachter, Mittelstadt, and Russell}]{wachter2017counterfactual}
Sandra Wachter, Brent Mittelstadt, and Chris Russell. 2018.
\newblock Counterfactual explanations without opening the black box: Automated decisions and the gdpr.
\newblock \emph{Harvard Journal of Law \& Technology}, 31(2):841--887.

\bibitem[{Webb(2002)}]{webb2002depth}
Norman~L Webb. 2002.
\newblock Depth-of-knowledge levels for four content areas.
\newblock \url{https://ossucurr.pbworks.com/w/file/fetch/49691156/Norm\%20web\%20dok\%20by\%20subject\%20area.pdf}.
\newblock Accessed: 2023-06-14.

\bibitem[{Wei et~al.(2022{\natexlab{a}})Wei, Tay, Bommasani, Raffel, Zoph, Borgeaud, Yogatama, Bosma, Zhou, Metzler et~al.}]{wei2022emergent}
Jason Wei, Yi~Tay, Rishi Bommasani, Colin Raffel, Barret Zoph, Sebastian Borgeaud, Dani Yogatama, Maarten Bosma, Denny Zhou, Donald Metzler, et~al. 2022{\natexlab{a}}.
\newblock Emergent abilities of large language models.
\newblock \emph{arXiv preprint arXiv:2206.07682}.

\bibitem[{Wei et~al.(2022{\natexlab{b}})Wei, Wang, Schuurmans, Bosma, Xia, Chi, Le, Zhou et~al.}]{cot}
Jason Wei, Xuezhi Wang, Dale Schuurmans, Maarten Bosma, Fei Xia, Ed~Chi, Quoc~V Le, Denny Zhou, et~al. 2022{\natexlab{b}}.
\newblock Chain-of-thought prompting elicits reasoning in large language models.
\newblock \emph{Advances in neural information processing systems}, 35:24824--24837.

\bibitem[{Wu et~al.(2021)Wu, Ribeiro, Heer, and Weld}]{wu2021polyjuice}
Tongshuang Wu, Marco~Tulio Ribeiro, Jeffrey Heer, and Daniel~S Weld. 2021.
\newblock Polyjuice: Generating counterfactuals for explaining, evaluating, and improving models.
\newblock In \emph{Proceedings of the 59th Annual Meeting of the Association for Computational Linguistics and the 11th International Joint Conference on Natural Language Processing}, pages 6707--6723.

\bibitem[{Xu et~al.(2023)Xu, Liu, Wu, and Wang}]{yao2021counterfactual}
Weizhi Xu, Qiang Liu, Shu Wu, and Liang Wang. 2023.
\newblock Counterfactual debiasing for fact verification.
\newblock In \emph{Proceedings of the 61st Annual Meeting of the Association for Computational Linguistics (Volume 1: Long Papers)}, pages 6777--6789.

\end{thebibliography}

\newpage
\onecolumn
\appendix

\section{Appendix}
\label{sec:appendix}
\subsection*{Examples of MalAlgoQA dataset}
\label{app:example-1}
We provide five math examples, one per content classifications, of question, choices and their corresponding rationales in Table~\ref{tab:math_example-1} and~\ref{tab:math_example-2}.
\begin{table*}[t]
\centering
\begin{minipage}[t]{1\textwidth}
\begin{lstlisting}[mathescape=true,basicstyle=\ttfamily\smalltonormalsize]
$\textbf{Math Question Set Example 1}$: 

Content Classification: Number & Operation
            
Question: What number is subtracted from 1,000 to result in a difference of 421?

Answer and Rationales:
$\colorbox{lightyellow}{A: 421}$
Rationale A: Selects the result of subtraction.
$\colorbox{lightyellow}{B: 579}$
Rationale B: 1000 - 421 = 579
$\colorbox{lightyellow}{C: 621}$
Rationale C: Rounded the hundreds place and added the tens and ones place. 1000 - 400 = 600; 600 + 21 = 621
$\colorbox{lightyellow}{D: 689}$
Rationale D: Rounded in hundreds, tens and ones place. 1000 - 400 = 600; 100 - 20 = 80; 10 - 1 = 9; 600 + 80 + 9 = 689

$\textbf{Math Question Set Example 2}$: 

Content Classification: Algebra
            
Question: Anne bought a calculator that cost 30. She received 10% off her purchase and then was charged 6% tax. What was the total amount that Anne paid?

Answer and Rationales:
$\colorbox{lightyellow}{A: 21.20}$
Rationale A: Subtracted 10 from 30, then added 6%.
$\colorbox{lightyellow}{B: 25.38}$
Rationale B: Subtracted 10% of the price from 30 then subtracted 6%.
$\colorbox{lightyellow}{C: 28.62}$
Rationale C: Subtracted 10% of the price from 30 and then added 6%.
$\colorbox{lightyellow}{D: 28.80}$
Rationale D: Calculated tax based on original price, then subtracted the 3.

$\textbf{Math Question Set Example 3}$: 

Content Classification: Geometry and Measurement
            
Question: Which side lengths form a right triangle?

Answer and Rationales:
$\colorbox{lightyellow}{A: 2 cm, 4 cm, 8 cm}$
Rationale A: Multiplied the 2 smaller sides to get the longest side.
$\colorbox{lightyellow}{B: 4 cm, 5 cm, 6 cm}$
Rationale B: Selected side with constant difference between the sides.
$\colorbox{lightyellow}{C: 5 cm, 12 cm, 13 cm}$
Rationale C: 5^2 + 12^2 = 13^2
$\colorbox{lightyellow}{D: 9 cm, 16 cm, 25 cm}$
Rationale D: Added the 2 smaller sides to get the longest side.
	
\end{lstlisting}
\end{minipage}
\hfill
\caption{Examples of Mathematics question set. }
\label{tab:math_example-1}
\end{table*}

\begin{table*}[t]
\centering
\begin{minipage}[t]{1\textwidth}
\begin{lstlisting}[mathescape=true,basicstyle=\ttfamily\smalltonormalsize]

$\textbf{Math Question Set Example 4}$: 

Content Classification: Data Analysis
            
Question: Mrs. Castillo recorded the number of students in each grade 5 classroom.
            24  28  25  24  29
What is the mean of the data?

Answer and Rationales:
$\colorbox{lightyellow}{A: 24}$
Rationale A: Chose the mode.
$\colorbox{lightyellow}{B: 25}$
Rationale B: Chose the median. {24, 24, 25, 28, 29}
$\colorbox{lightyellow}{C: 26}$
Rationale C: (24 + 28 + 25 + 24 + 29) / 5 = 130/5 = 26
$\colorbox{lightyellow}{D: 29}$
Rationale D: Chose the greatest number.

$\textbf{Math Question Set Example 5}$: 

Content Classification: Probability
            
Question: Ms. Collier had a deck of cards. There were stars on 1/4 of the cards in the deck. After randomly picking a card and returning it to the deck 100 times, the expected result and relative frequency of picking a card with a star were equal. In the first 50 cards she picked, she got a star 10 times. How many times did Ms. Collier get a star in the second 50 cards she picked?

Answer and Rationales:
$\colorbox{lightyellow}{A: 10}$
Rationale A: Chose the number of times a star is picked in the 1st 50 cards drawn.
$\colorbox{lightyellow}{B: 13}$
Rationale B: Calculated 50 * 1/4 = 12.5 and then rounded to 13.
$\colorbox{lightyellow}{C: 15}$
Rationale C: Correct. 100 * 1/4 = 25 cards w/ stars. 25 - 10 = 15
$\colorbox{lightyellow}{D: 25}$
Rationale D: Calculated 100 * 1/4 = 25 cards w/ stars.

	
\end{lstlisting}
\end{minipage}
\hfill
\caption{Examples of Mathematics question set. }
\label{tab:math_example-2}
\end{table*}






\end{document}